%% file: main.tex
\documentclass{svproc}
\usepackage[T1]{fontenc}
\def\doi#1{\href{https://doi.org/\detokenize{#1}}{\url{https://doi.org/\detokenize{#1}}}}
\usepackage[title,toc,titletoc,header]{appendix}
\usepackage{graphicx}
\usepackage[hidelinks]{hyperref}

\usepackage{listings}
\usepackage{xspace}
\usepackage{amsmath} 
\usepackage{amssymb}  
\usepackage{xcolor}
\usepackage{algorithm, algpseudocode}
\usepackage{cleveref}
\usepackage{array}
\usepackage{breakcites}

\include{shortcut}

\DeclareMathOperator*{\argmax}{arg\,max}

\newcounter{cmt}

\makeatletter
\renewcommand\section{\@startsection{section}{1}{\z@}%
                       {-12\p@ \@plus -2\p@ \@minus -2\p@}%
                       {6\p@ \@plus 2\p@ \@minus 2\p@}%
                       {\normalfont\large\bfseries\boldmath
                        \rightskip=\z@ \@plus 8em\pretolerance=10000 }}
\renewcommand\subsection{\@startsection{subsection}{2}{\z@}%
                       {-9\p@ \@plus -2\p@ \@minus -2\p@}%
                       {4\p@ \@plus 2\p@ \@minus 2\p@}%
                       {\normalfont\normalsize\bfseries\boldmath
                        \rightskip=\z@ \@plus 8em\pretolerance=10000 }}
\renewcommand\subsubsection{\@startsection{subsubsection}{3}{\z@}%
                       {-3\p@ \@plus -1\p@ \@minus -1\p@}%
                       {-0.5em \@plus -0.22em \@minus -0.1em}%
                       {\normalfont\normalsize\bfseries\boldmath}}
\renewcommand\paragraph{\@startsection{paragraph}{4}{\z@}%
                       {-3\p@ \@plus -1\p@ \@minus -1\p@}%
                       {-0.5em \@plus -0.22em \@minus -0.1em}%
                       {\normalfont\normalsize\itshape}}
\makeatother

\begin{document}
\mainmatter              
\title{Adaptive Discretization using Voronoi Trees for Continuous-Action POMDPs} 
\titlerunning{ADVT for Continuous-Action POMDPs}
%
\author{Marcus Hoerger\inst{1}
\and
Hanna Kurniawati\inst{2}
\and
Dirk Kroese\inst{1}
\and
Nan Ye\inst{1}}
\authorrunning{M. Hoerger et al.}
%
\institute{School of Mathematics \& Physics, The University of Queensland, Australia \\
\email{m.hoerger@uq.edu.au, kroese@maths.uq.edu.au, nan.ye@uq.edu.au}\\
\and
School of Computing, Australian National University, Australia \\
\email{hanna.kurniawati@anu.edu.au}}
\maketitle              
\begin{abstract}
	Solving Partially Observable Markov Decision Processes \\ (POMDPs) with continuous actions is challenging, particularly for high-dimensional action spaces. To alleviate this difficulty, we propose a new sampling-based online POMDP solver, called \emph{\solver (\solverAbbr)}. It uses Monte Carlo Tree Search in combination with an adaptive discretization of the action space as well as optimistic optimization to efficiently sample high-dimensional continuous action spaces and compute the best action to perform. 
	Specifically, we adaptively discretize the action space for each sampled belief using a
	hierarchical partition which we call a \emph{Voronoi tree}. 
	A Voronoi tree is a Binary Space Partitioning (BSP) that implicitly maintains
	the partition of a cell as the Voronoi diagram of two points sampled from the
	cell.
	This partitioning strategy keeps the cost of partitioning and estimating the size of each cell low, even in high-dimensional spaces where many sampled points are required to cover the space well. \solverAbbr uses the estimated sizes of the cells to form an upper-confidence bound of the action values of the cell, and in turn uses the upper-confidence bound to guide the Monte Carlo Tree Search expansion and further discretization of the action space. This strategy enables \solverAbbr to better exploit local information in the action space, leading to an action space discretization that is more adaptive, and hence more efficient in computing good POMDP solutions, compared to existing solvers. Experiments on simulations of four types of benchmark problems indicate that \solverAbbr outperforms and scales substantially better to high-dimensional continuous action spaces, compared to state-of-the-art continuous action POMDP solvers. 	
\end{abstract}
\section{Introduction}
\label{s:Intro}

Planning in scenarios with non-deterministic action effects 
and partial observability is an essential, yet
challenging problem for autonomous robots. The Partially Observable Markov
Decision Process (POMDP)\ccite{kaelbling1998planning,Sondik:71} is a general principled framework for
such planning problems. POMDPs lift the planning problem from the state
space to the \textit{belief space} ---that is, the set of all probability
distributions over the state space. By doing so, POMDPs enable robots to
systematically account for uncertainty caused by stochastic actions and
incomplete or noisy observations in computing the optimal strategy. Although
computing the optimal strategy exactly is intractable in general\ccite{papadimitriou1987complexity}, the past two decades have seen a surge
of sampling-based POMDP solvers (reviewed in\ccite{Kur22:Partially}) that trade optimality with approximate
optimality for computational tractability, enabling POMDPs to become practical for a variety of realistic robotics problems.

Despite these advances, POMDPs with high-dimensional continuous action spaces remain a challenge. 
Recent solvers for continuous-action POMDPs\ccite{fischer2020information,mern2021bayesian,seiler2015online,sunberg2018online} are generally online ---that is, planning and execution are interleaved--- and exploit Monte Carlo Tree Search (MCTS) to find the best action among a finite representative subset of the action space. MCTS interleaves guided belief space sampling, value estimation and action subset refinement to incrementally improve the possibility that the selected subset of actions contains the best action.
They generally use UCB1\ccite{Aue02:Finite} to guide belief space sampling and Monte Carlo backup for value estimation, but differ in the action subset refinement. 


Several approaches use the Progressive Widening strategy\ccite{couetoux2011continuous} to continuously add new randomly sampled actions once current actions have been sufficiently explored. Examples include POMCPOW\ccite{sunberg2018online} and IPFT\ccite{fischer2020information}. More recent algorithms combine Progressive Widening with more informed methods for adding new actions: VOMCPOW\ccite{lim2020voronoi} uses Voronoi Optimistic Optimization\ccite{kim2020monte} and BOMCP\ccite{mern2021bayesian} uses Bayesian optimization. All of these solvers use UCT-style simulations and Monte Carlo backups. An early line of work, GPS-ABT\ccite{seiler2015online}, takes a different approach: It uses Generalized Pattern Search to iteratively select an action subset that is more likely to contain the best action and add it to the set of candidate actions. GPS-ABT uses UCT-style simulations and Bellman backup (following the implementation of ABT\ccite{opptURL,tapirURL}), though the distinction between Monte Carlo and Bellman backup was not  clarified nor explored.  All of these solvers have been successful in finding good solutions to POMDPs with continuous action spaces, though for a relatively low ($\le 4)$ dimension. 

To compute good strategies for POMDPs with high-dimensional action spaces, we propose a new online POMDP solver, called \solver (\solverAbbr).
\solverAbbr is motivated by the observation that in many continuous action POMDPs for robotics problems, the distance between two actions can often be used as an indication of how similar their values are.
Using this observation, \solverAbbr assumes that the action value for a belief is Lipschitz continuous in the action space and proposes a new action space discretization mechanism called \emph{Voronoi tree}. 
A Voronoi tree represents a hierarchical partition of the action space for a single sampled belief. It follows the structure of a Binary Space Partitioning (BSP) tree, but each partitioning hyper-plane is only implicitly maintained and computed based on the Voronoi diagram of a pair of sampled actions. 
This strategy enables \solverAbbr to keep a low computational cost for partitioning and estimating the diameters of the cells, even in high-dimensional action spaces. 

\solverAbbr uses the estimated diameters of the cells in a Voronoi tree to guide
belief space sampling and Voronoi tree refinement in two ways. First, they help decide if a cell needs further refinement, which in turn helps \solverAbbr to avoid unnecessarily small partitioning of non-promising action space regions. 
Our refinement rule, together with the hierarchical partitioning, results in a partitioning that is much more adaptive to the spatial locations of the sampled actions, 
compared to state-of-the-art methods\ccite{bubeck2011x,lim2020voronoi,mansley2011sample,sunberg2018online,valko2013stochastic}. 
The second use of the information on the diameters of the cells is in the action sampling
strategy used in MCTS.
Instead of using UCB1\ccite{Aue02:Finite}, \solverAbbr adopts a cell-diameter-aware upper-confidence bound\ccite{wang2020towards}, which uses the diameter of 
a cell to estimate the upper-confidence bound of the value of \emph{all} actions within the cell. 
The above strategies imply that \solverAbbr uses local information to help construct a highly adaptive discretization of the continuous action space and guide the search for the optimal action. Finally, \solverAbbr applies stochastic Bellman backups, rather than the typical Monte Carlo backup.


Experimental results on a variety of benchmark problems with increasing dimension (up to 12-D) of the action space indicate that \solverAbbr substantially outperforms state-of-the-art methods\ccite{lim2020voronoi,sunberg2018online}. \commNew{Our C++ implementation of ADVT is available at \url{https://github.com/hoergems/ADVT}.}

\section{Background and Related Work}

A POMDP provides a general mathematical framework for sequential decision making
under uncertainty.
Formally, it is an 8-tuple 
$\langle \stSpace, \actSpace, \obsSpace, \transF, \obsF, \rewFunc, \bel_{0}, \gamma \rangle$.
The robot is initially in a hidden state $s_{0} \in \stSpace$. 
This uncertainty is represented by an initial belief $\bel_0$, which is a
probability distribution on the state space $\stSpace$.
At each step $t \ge 0$, the robot executes an action $\act_{t} \in \actSpace$
according to some policy $\pol$.
It transitions to a next state $\st_{t+1} \in \stSpace$ according to 
$\st_{t+1}\sim\transF(\st_{t}, \act_{t}, \st_{t+1}) = p(\st_{t+1} | \st_{t}, \act_{t})$. In this paper, the function \transF is a probability density function, as both state and actions spaces are continuous. 
The robot does not know the state $\st_{t+1}$ exactly, but perceives an observation $\obs_{t} \in\obsSpace$ with probability 
$\obsF(\st_{t+1}, \act_{t}, \obs_{t}) = p(\obs_{t} | \st_{t+1}, \act_{t})$.
In addition, it receives an immediate reward $r_{t} = R(\st_{t}, \act_{t}) \in \reals$.
The robot's goal is to find a policy $\pol$ that maximizes the expected total
discounted reward; that is, the value
$V_{\pol}(\bel_{0}) = \sum_{t=0}^{\infty}\gamma^t \expect{r_t | \bel_{0}}$,
where the discount factor $0 < \gamma < 1$ ensures that $V_{\pol}(\bel)$ is
finite and well-defined.

The robot's decision space is the set $\Pi$ of policies defined as mappings
from beliefs to actions. 
The POMDP solution is then the optimal policy, denoted as \optPol and defined as  $\optPol = \argmax_{\pol \in \Pi}
V_{\pol}(\bel)$. 
In designing solvers, it is often convenient to work with the action value or $Q$-value 
$Q(\bel, \act) 
= R(\bel, \act) 
+ 
\gamma \max_{\pol \in \Pi} \mathbb{E}_{\obs\in\obsSpace}[V_{\pol}(\bel_{\act}^{\obs}) | \bel]$,
where $R(\bel, \act) = \int_{\st\in\stSpace}\bel(\st)R(\st, \act)\del\st$ is the expected reward of executing action $\act$ at belief $\bel$, while  $\bel_{\act}^{\obs} = \tau(\bel, \act, \obs)$ is the updated robot's belief estimate after it performs action ${\act \in \actSpace}$  while at belief \bel, and subsequently perceives observation $\obs \in \obsSpace$. The optimal value function is then 
$V^*(\bel) = \max_{\act\in\actSpace} Q(\bel, \act)$. A more elaborate explanation is available in\ccite{kaelbling1998planning}.

Belief trees are convenient data structures to find good approximations  to the optimal solutions via sampling-based approaches, which has been shown to significantly improve the scalability of POMDP solving\ccite{Kur22:Partially}. Each node in a belief tree represents a sampled belief. It has outgoing edges labeled by actions, and each action edge is followed by outgoing edges labeled by observations and leading to updated belief nodes. Na\"ively, bottom-up dynamic programming can be applied to a truncated belief tree to obtain a near-optimal policy, but many scalable POMDP solvers use more sophisticated strategies to construct a compact belief tree, from which a close-to-optimal policy can be computed efficiently. \solverAbbr uses such a sampling-based approach and belief tree representation too.

Various efficient sampling-based offline and online POMDP solvers have been developed for increasingly complex discrete and continuous POMDPs in the last two decades. Offline solvers (e.g.,\ccite{bai2014integrated,kurniawati2011motion,Kurniawati08sarsop:efficient,Pin03:Point,Smi05:Point}) compute an optimal policy for all beliefs first before deploying them for execution. In contrast, online solvers (e.g.,\ccite{kurniawati2016online,silver2010monte,somani2013despot}) aim to further scale to larger and more complex problems by interleaving planning and execution, and focusing on computing an optimal action for only the current belief during planning. For scalability purposes, \solverAbbr follows the online solving approach.

Some online solvers have been designed for continuous POMDPs. In addition to the general solvers discussed in \sref{s:Intro}, some solvers\ccite{agha2011firm,sun2015high,van2011lqg,van2012motion} restrict beliefs to be Gaussian and use LQG to compute the best action. This strategy generally performs well in high-dimensional action spaces. However, they tend to perform poorly in problems with large uncertainties\ccite{hoerger2020linearization}.


Last but not least, hierarchical rectangular partitions have been commonly applied to solve continuous action bandits and MDPs (the fully observed version of POMDPs), such as HOO\ccite{bubeck2011x} and HOOT\ccite{mansley2011sample}. However, the partitions used in these algorithms are typically predefined, which are less adaptive than Voronoi-based partitions constructed dynamically during the search. On the other hand, Voronoi partitions have been proposed in VOOT\ccite{kim2020monte} and VOMCPOW\ccite{lim2020voronoi}. However, their partitions are based on the Voronoi diagram of all sampled actions, which makes the computation of cell diameters and sampling relatively complex in high-dimensional action spaces. 
\solverAbbr is computationally efficient, just like hierarchical rectangular partitions, and yet adaptive, just like the Voronoi partitions, getting the best of both worlds.

\section{\solverAbbr: Overview}

In this paper, we consider a POMDP $\pomdpTuple = \langle \stSpace, \actSpace, \obsSpace, \transF, \obsF, \rewFunc, \bel_{0}, \gamma \rangle$, where the action space \actSpace is continuous and embedded in a bounded metric space with distance function $d$. Typically, we define the metric space to be a $D$-dimensional bounded Euclidean space, though \solverAbbr can also be used with other types of bounded metric spaces. We also consider the state space \stSpace to be continuous or discrete, while we assume the observation space \obsSpace to be discrete. 

\solverAbbr assumes that the $Q$-value is Lipschitz continuous in the action space; that is, for any belief $\bel \in \belSpace$, there exists a Lipschitz constant $L_b$ such that for any actions $\act, \act' \in \actSpace$, we have $\left |Q(\bel, \act) - Q(\bel, \act') \right | \leq L_b\ d(\act, \act')$. 
Since generally we do not know a tight Lipschitz constant, in the implementation, \solverAbbr uses the same Lipschitz constant $L$ for all beliefs in \belSpace, as discussed in \sref{ssec:actSelection}.

\solverAbbr is an anytime online solver for POMDPs. 
It interleaves belief space sampling and action space sampling to find the best action to perform from the current belief $\bel \in \belSpace$. 
The sampled beliefs are maintained in a belief tree, denoted as \belTree, while the sampled actions $\actSpace(b)$ 
for a belief $b$ are maintained in a Voronoi tree, denoted as $\VT(b)$. 
The Voronoi trees form part of the belief tree in \solverAbbr:
they determine the sampled action branches for the belief nodes.
\aref{alg:mcts} presents the overall algorithm of \solverAbbr, with details in the sections below.


\begin{algorithm}[tbp]
\caption{\textproc{\solverAbbr}(Initial belief $\bel_0$)}\label{alg:mcts}
\begin{algorithmic}[1]
\State $\bel = \bel_0$; 
\State $\belTree = $\ initializeBeliefTree($\bel$) 
\State $\VT(\bel) = $ Initialize Voronoi tree for belief \bel
\State isTerminal $=$\ False
\While{isTerminal is False}
    \While{planning budget not exceeded}
        \State $(\episode, \bel_c) = $\ \textproc{SampleEpisode}(\belTree, \bel)\Comment{\aref{alg:sampleEpisode}}        
        \For{$i = \left |\episode \right | - 1$ to 1}
               \State $(\st, \act, \obs, r) = \episode_i$
		\State \textproc{Backup}(\belTree, $\bel_c$, \act, $r$) \Comment{\aref{alg:backup}}
		\State $\bel_c = $\ Parent node of $\bel_c$ in \belTree
		\State \textproc{RefineVoronoiTree}($\VT(\bel_c)$, \act) \Comment{\aref{alg:updateAction}}		
        \EndFor
    \EndWhile     
    \State $\act =$\ Get best action in \belTree from \bel 
    \State $(\obs$,\ isTerminal$) =$\ Execute \act    
    \State $\bel = \tau(\bel, \act, \obs)$     
\EndWhile
\end{algorithmic}
\end{algorithm}

\section{\solverAbbr: Construction of the Belief Tree}
\label{ssec:beliefTree}

\begin{algorithm}[htbp]
\caption{\textproc{SampleEpisode}(Belief tree \belTree, Belief node $\bel_c$)}\label{alg:sampleEpisode}
\begin{algorithmic}[1]
\State $\episode = $\ init episode; $\bel = \bel_c$; $\st = $\ sample a state from \bel; newBelief $=$ False
\While{newBelief is False and \st not terminal}
    \State $\actSpace(\bel) = $\ Set of candidate actions associated to the leaf-nodes of $\VT(\bel)$ 
    \State $\act = \argmax_{\act_k\in\actSpace(\bel)} U(\bel, \act_k)$
\Comment{\eref{eq:u_value}}
  \State $(\stp, \obs, r) = G(\st, \act)$ \Comment{Generative model}  
  \State append $(\st, \act, \obs, r)$ to $\episode$
  \State $N(\bel, \act) = N(\bel, \act) + 1; N(\bel) = N(\bel) + 1$
  \State $\st = \stp$
  \State $\bel = $ child node of \bel via edge $(\act, \obs)$ (If no such child exists, $\bel = null$)
  \If {$\bel = null$}
  	\State $\bel = $ Create a new belief node as a child of \bel via edge $(\act, \obs)$
  	\State $\VT(\bel) = $ Initialize Voronoi tree for belief \bel
  	\State newBelief = True ;  $N(\bel) = 0$
  \EndIf
\EndWhile
\State $r = 0$
\If{newBelief is True}
	\State $h = $\ calculateRolloutHeuristic(\st, \bel)
  	\State Initialize $\widehat{V}^*(\bel)$ with $h$
\EndIf
\State insert $(s, -, -, r)$ to $\episode$
\State \Return $(\episode, \bel)$
\end{algorithmic}
\end{algorithm}

The belief tree \belTree is a tree whose nodes represent beliefs and the edges are associated with action--observation pairs $(\act, \obs)$, where $\act \in \actSpace$ and $\obs \in \obsSpace$. A node $\bel'$ is a child of node $\bel$ via edge $(\act, \obs$) if and only if $\bel' = \tau(\bel, \act, \obs)$.


To construct the belief tree \belTree, \solverAbbr interleaves the iterative
select-expand-simulate-backup operations used in many MCTS algorithms with
adaptive discretization.
Each node $b$ in \belTree is associated with a finite action set
$\actSpace(\bel) \subset \actSpace$ which is adaptively refined.
At each iteration, it first \emph{selects} a path starting from the root by sampling
an episode 
$\st_{0}, \act_{0}, \obs_{0}, r_{0}, \st_{1}, \act_{1}, \obs_{1}, r_{1},
\ldots$ as follows:
Set the current node $\bel$ as the root node, and sample $\st_{0}$ from $\bel$;
at each step $i \ge 0$, choose an action $\act_{i} \in \actSpace(\bel)$ for
$\bel$ using an action selection strategy (discussed in \sref{ssec:actSelection}), execute $\act_{i}$ from state $\st_i$ to obtain $\obs_{i}, r_{i}$ and $\st_{i+1}$ as the observation, the immediate reward and
the next state by simulating the dynamics and observation model, and finally
update $\bel$ to $\bel$'s child node via $(\act_{i}, \obs_{i})$.
The process terminates when encountering a terminal state or when the child node
does not exist; in the latter case, the tree is \emph{expanded} by adding a new node,
and a rollout policy is \emph{simulated} to provide an estimated value for the new
node.
In either case, \emph{backup} operations are performed to update the estimated
values for all encountered actions.
In addition, $\actSpace({\bel})$ is associated with a Voronoi tree $\VT(\bel)$ in
\solverAbbr, and it is refined as needed for each encountered belief node $\bel$. 
\aref{alg:sampleEpisode} presents the pseudo-code for constructing \belTree, while the backup operation and refinement of $\actSpace(\bel)$ are discussed in \sref{ssec:backup} and \sref{ssec:voronoiForest}, respectively.

\subsection{Action Selection Strategy}
\label{ssec:actSelection}

In contrast to many existing online solvers, which use UCB1 to select the action to expand a node \bel of \belTree, \solverAbbr treats the action selection problem as a continuous-arm bandit problem. Specifically, it selects an action from the set of candidate actions $\actSpace(\bel)$ according to\ccite{wang2020towards}
\begin{align}
	\act^{*} &= \argmax_{\act \in\actSpace(\bel)} U(\bel, \act), \qquad\text{with} 
	    \label{eq:action_selection} \\
    U(\bel, \act) &= \widehat{Q}(\bel, \act) + C\sqrt{\frac{\log N(\bel)}{N(\bel, \act)}} + L\ \mathrm{diam}(\cell),
        \label{eq:u_value}
\end{align}
where $N(\bel)$ is the number of times node \bel has been visited so far,
\commNew{$\cell \subseteq \actSpace$ is the unique leaf cell containing $\act$ in
$\VT(\bel)$ (see \Cref{ssec:voronoiForest} for details on the Voronoi tree)},
and $\mathrm{diam}(\cell) = \sup_{\act, \act' \in \cell} d(\act, \act')$ is the
diameter of $\cell$ wrt distance metric $d$.
The constant $C$ is an exploration constant, where larger values of $C$ encourage exploration. 
In case $N(\bel, \act) = 0$, we set $U(\bel, \act) = \infty$. 
With the Lipschitz continuity assumption, the value $U(\bel, \act)$ can be seen as an upper-confidence bound for
the maximum possible Q-value $\max_{\act' \in \cell} Q(\bel, \act')$ within $P$, as follows:
$\widehat{Q}(\bel, \act) + C\sqrt{\frac{\log N(\bel)}{N(\bel, \act)}}$ is 
the standard UCB1 bound for the Q-value $Q(b, \act)$, and whenever this 
upper bounds $Q(\bel, \act)$, we have 
$U(\bel, \act) \ge Q(\bel, \act')$ for any $\act' \in \cell$, because
$U(\bel, \act) 
\ge Q(\bel, \act) + L\ \mathrm{diam}(\cell) 
\ge Q(\bel, \act')$, where the last inequality holds due to the Lipschitz assumption.
Since $L$ is unknown, we try different values of $L$ in our experiments and choose the best.

\subsection{Backup}
\label{ssec:backup}


\begin{algorithm}[htbp]
\caption{\textproc{Backup}(Belief tree \belTree, Belief node $\bel'$, Action \act, Reward $r$)}\label{alg:backup}
\begin{algorithmic}[1]
        \State $\bel = $\ Parent node of $\bel'$ in \belTree
	\State {$\widehat{Q}(\bel, \act) = \widehat{Q}(\bel, \act) + N(\bel, \act)^{-1}(r + \gamma \widehat{V}^*(\bel') - \widehat{Q}(\bel, \act))$}
	\State {$\widehat{V}^*(\bel) = \max_{\act\in\actSpace(\bel)} \widehat{Q}(\bel, \act)$}
\end{algorithmic}
\end{algorithm}
After sampling an episode, \solverAbbr updates the estimates $\widehat{Q}(\bel, \act)$ as well as the statistics $N(\bel)$ and $N(\bel, \act)$ along the sequence of beliefs visited by the episode. To update $\widehat{Q}(\bel, \act)$, we use a stochastic version of the Bellman backup (\aref{alg:backup}): Suppose $r$ is the immediate reward sampled by the episode after selecting \act from \bel. We then update $\widehat{Q}(\bel, \act)$ according to 
$\widehat{Q}(\bel, \act) = \widehat{Q}(\bel, \act) + N(\bel, \act)^{-1}(r + \gamma \widehat{V}^*(\bel') - \widehat{Q}(\bel, \act))$, 
where $\bel'$ is the child of \bel in the belief tree \belTree via edge $(\act, \obs)$; \ie, the belief we arrived at after performing action $\act \in \actSpace$ and perceiving observation ${\obs \in \obsSpace}$ from \bel. Note that this rule is in contrast to POMCP, POMCPOW and VOMCPOW, where the $Q$-value estimates are updated via Monte-Carlo backup. This update rule is akin to the rule used in $Q$-Learning\ccite{watkins1992q} and was implemented in the ABT software\ccite{opptURL,tapirURL}, though never explicitly compared with Monte Carlo backup. The above update rule helps \solverAbbr to focus its search on promising parts of the belief tree, particularly for problems where good rewards are sparse.

\section{\solverAbbr: Construction and Refinement of Voronoi Trees}
\label{ssec:voronoiForest}


For each belief node $\bel$ in the belief tree, its Voronoi tree $\VT(\bel)$ is
a BSP tree for $\actSpace$. Each node in $\VT(\bel)$ consists of a pair $(\act,
\cell)$ with $P \subseteq \actSpace$ and $a$ the representative action of $P$,
and each non-leaf node is partitioned into two child nodes.
The partition of each cell in a Voronoi tree is a Voronoi diagram for two
actions sampled from the cell.


\begin{algorithm}[!htbp]
\caption{\textproc{RefineVoronoiTree}(Voronoi tree $\VT(\bel)$, Action \act)}\label{alg:updateAction}
\begin{algorithmic}[1]
        \State $(a, \cell) = $\ leaf node of $\VT(\bel)$ with its action component being $\act$
        \If{$C_{r}N(\bel, \act) \geq 1/\mathrm{diam}(\cell)^2$}
            \State $\act'$ = sample from $\cell$
			\State $(\cell_{1}, \cell_{2}) = $\ Child cells of $\cell$ induced by \act and $\act'$
			\State Compute diameters of $\cell_{1}$ and $\cell_{2}$
			\State Add $(\act, \cell_{1})$ and $(\act', \cell')$ as $(\act, \cell)$'s children
        \EndIf	
\end{algorithmic}
\end{algorithm}

\begin{figure*}
\centering
\small
\includegraphics[width=0.8\textwidth]{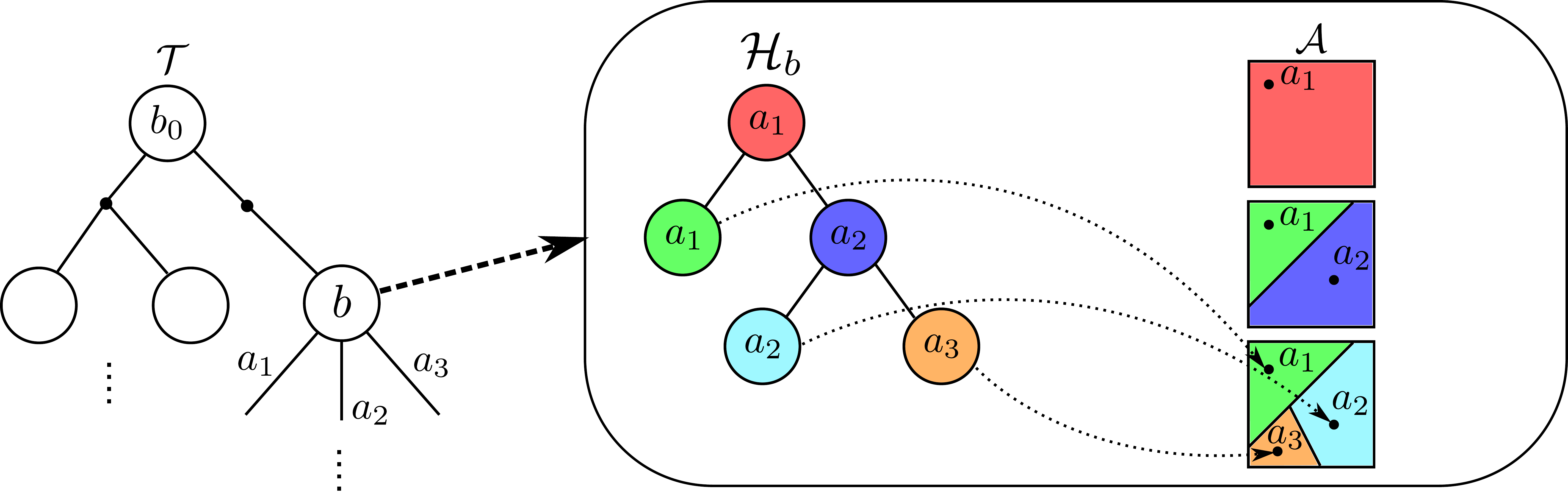}
\caption{Illustration of the relation between a belief tree \belTree (left), the Voronoi tree $\VT(\bel)$ associated to belief \bel (middle) and the partition of the action space induced by the Voronoi tree (right).}
\label{f:Partitioning}
\end{figure*}

To construct $\VT(\bel)$, \solverAbbr first samples an action $\act$ uniformly at random from the action space \actSpace, and sets the pair $(\act, \actSpace)$ as the root of $\VT(\bel)$. When \solverAbbr decides to expand the node $(\act, \cell)$, it first samples an action $\act'$ uniformly at random from $\cell \subseteq \actSpace$. \solverAbbr then implicitly constructs the Voronoi diagram between $\act$ and $\act_i$ within the cell $\cell$, splitting it into two regions: One is $\cell_1$, representing the set of all actions $\act'' \in \cell$ for which the distance $d(\act'', \act) < d(\act'', \act')$, and the other is $\cell_2 = \cell \backslash \cell_1$. The nodes $(\act, \cell_1)$ and $(\act', \cell_2)$ are then inserted as children of $(\act, \cell)$ in $\VT(\bel)$. 
The leaf nodes of $\VT(\bel)$ form the partition of the action space \actSpace used by belief \bel, while the finite action subset $\actSpace(\bel) \subset \actSpace$ used to find the best action from \bel is the set of actions associated with the leaves of $\VT(\bel)$.  \fref{f:Partitioning} illustrates the relationship between a belief, the Voronoi tree $\VT(\bel)$ and the partition of \actSpace. 



Three key details in constructing the tree $\VT(\bel)$ are elaborated below. \sref{ssec:refineCell} describes how  \solverAbbr decides which node of $\VT(\bel)$ to expand. \commNew{\sref{ssec:estDiam} presents how the diameter of a Voronoi cell is estimated}. Finally, \sref{ssec:samplingCells} describes how uniform sampling is performed efficiently within a cell.


\subsection{Refining the Partition}
\label{ssec:refineCell}

\solverAbbr decides how to refine the partitioning $\VT(\bel)$ in two steps. First, it selects a leaf node of $\VT(\bel)$ to be refined next. This step relies on the action selection strategy used for expanding the belief tree \belTree (\sref{ssec:actSelection}). The selected leaf node $(\act, \cell)$ of $\VT(\bel)$ is the unique leaf node with $\act$ chosen according to \Cref{eq:action_selection}

In the second step \solverAbbr decides if the cell \cell\ should indeed be refined, based on the quality of the estimate $\widehat{Q}(\bel, \act)$ and the variation of the $Q$-values for the actions contained in \cell. Specifically, \solverAbbr refines the cell \cell\ only when the following criteria is satisfied:
\begin{equation}\label{eq:refinement_condition}
    C_{r} N(\bel, \act) \geq \frac{1}{\mathrm{diam}(\cell)^2},
\end{equation}
where $C_{r}$ is an exploration constant and  $N(\bel, \act)$ is the number of times that $\act$ has been selected at
$\bel$, which provides a rough estimate on the quality of the $\widehat{Q}(\bel, \act)$ estimate. This criterion, which is inspired by\ccite{touati2020zooming}, limits the growth of the finite set of candidate actions $\actSpace(\bel)$ and ensures that a cell is only refined when its corresponding action has been played sufficiently often. Larger $C_r$ cause cells to be refined earlier, thereby encouraging exploration.

Our refinement strategy is highly adaptive, in the sense that we use local information (i.e., the size of the cells, induced by the sampled actions), to influence the choice of the cell to be partitioned and when the chosen cell is partitioned, and the geometries of our cells are dependent on the sampled actions. 
This strategy is in contrast to other hierarchical decompositions, such as those used in HOO and HOOT, where the cell that corresponds to an action is refined immediately after the action is selected for the first time, which generally means the $Q$-value of an action is estimated based only on a single play of the action, which is grossly insufficient for our problem.
In addition, our strategy is more adaptive than VOMCPOW in the sense that we use local information to decide when to refine the decomposition.

\subsection{Estimating the Voronoi Cell Diameters}
\label{ssec:estDiam}

\solverAbbr uses the diameters of the cells in the action selection strategy and the cell refinement rule, but efficiently computing the  diameters of the cells is computationally challenging in high-dimensional spaces.
We give an efficient approximation algorithm for computing the Voronoi cell diameters below.


Since the cells in $\VT(\bel)$ are only implicitly defined, we use a sampling-based approach to approximate a cell's diameter. Suppose we want to estimate the diameter of the cell \cell\ corresponding to the node $(\act, \cell)$ of the Voronoi tree $\VT(\bel)$. Then, we first sample a set of points ${\actSpace}_{\cell}(\bel)$ that approximately lie on the boundary of \cell. To sample a boundary point $\act_\cell \in {\actSpace}_{\cell}(\bel) $, we first sample a point $\alpha$ that lies on the sphere centered at $\act$ with diameter $\mathrm{diam}(\actSpace)$ -- which can be easily computed for our benchmark problems -- uniformly at random. 
The point $\alpha$ lies either on the boundary or outside of \cell. We then use the Bisection method\ccite{Burden2016numerical} with \act and $\alpha$ as the initial end-points, until the  two end-points are less than a small threshold $\epsilon$ away from each other, but one still lies inside \cell\ and the other outside \cell. The point that lies inside \cell\ is then a  boundary point $\act_\cell$. The diameter of a bounding sphere that encloses all the sampled boundary points in ${\actSpace}_{\cell}(\bel)$\ccite{Welzl91enclosing} is then an approximation of the diameter of \cell.


\subsection{Sampling from the Voronoi Cells}
\label{ssec:samplingCells}

To sample an action that is approximately uniformly distributed in a cell \cell, we use a simple Hit \& Run approach\ccite{smith1984efficient} that performs a random walk within \cell. 
Suppose \cell\ is the cell corresponding to the node $(\act, \cell)$ of the Voronoi tree $\VT(\bel)$. We first sample an action $\act_\cell$ on the boundary of \cell\ using the method described in \sref{ssec:estDiam}. Subsequently, we take a random step from \act in the direction towards $\act_\cell$, resulting in a new action $\act'\in\cell$. We then use $\act'$ as the starting point, and iteratively perform this process for $m$ steps, which gives us a point that is approximately uniformly distributed in \cell.


\section{Experiments and Results}

We evaluated \solverAbbr on 4 robotics tasks, formulated as continuous-action POMDPs.

\subsection{Problem Scenarios}

\begin{figure}[htb]
\centering
\begin{tabular}{c@{\hskip5pt}c@{\hskip5pt}c}
\includegraphics[height=0.25\textwidth]{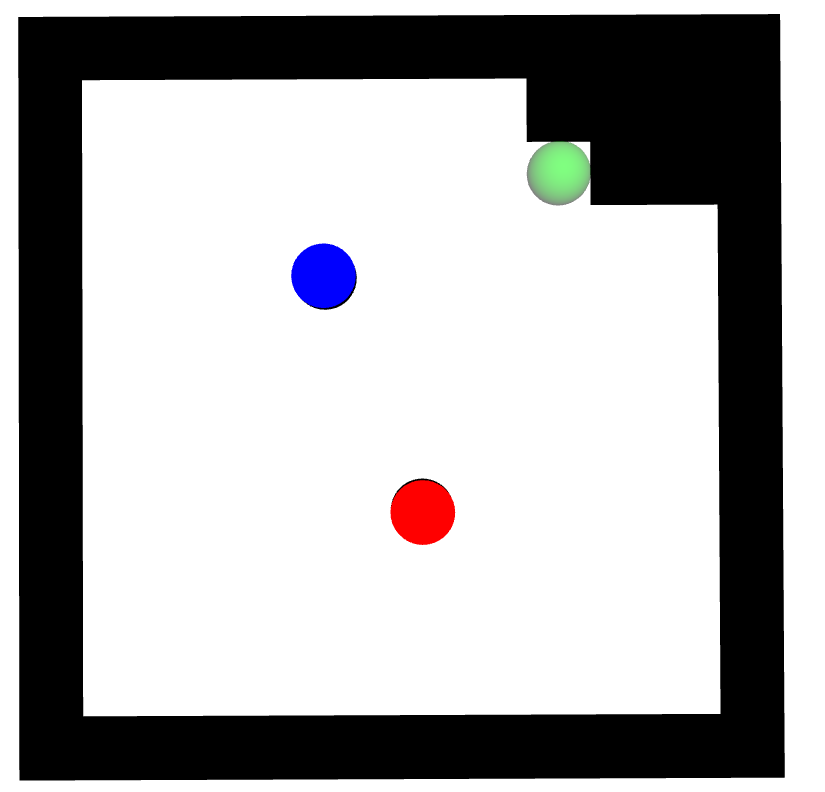} &
\includegraphics[height=0.25\textwidth]{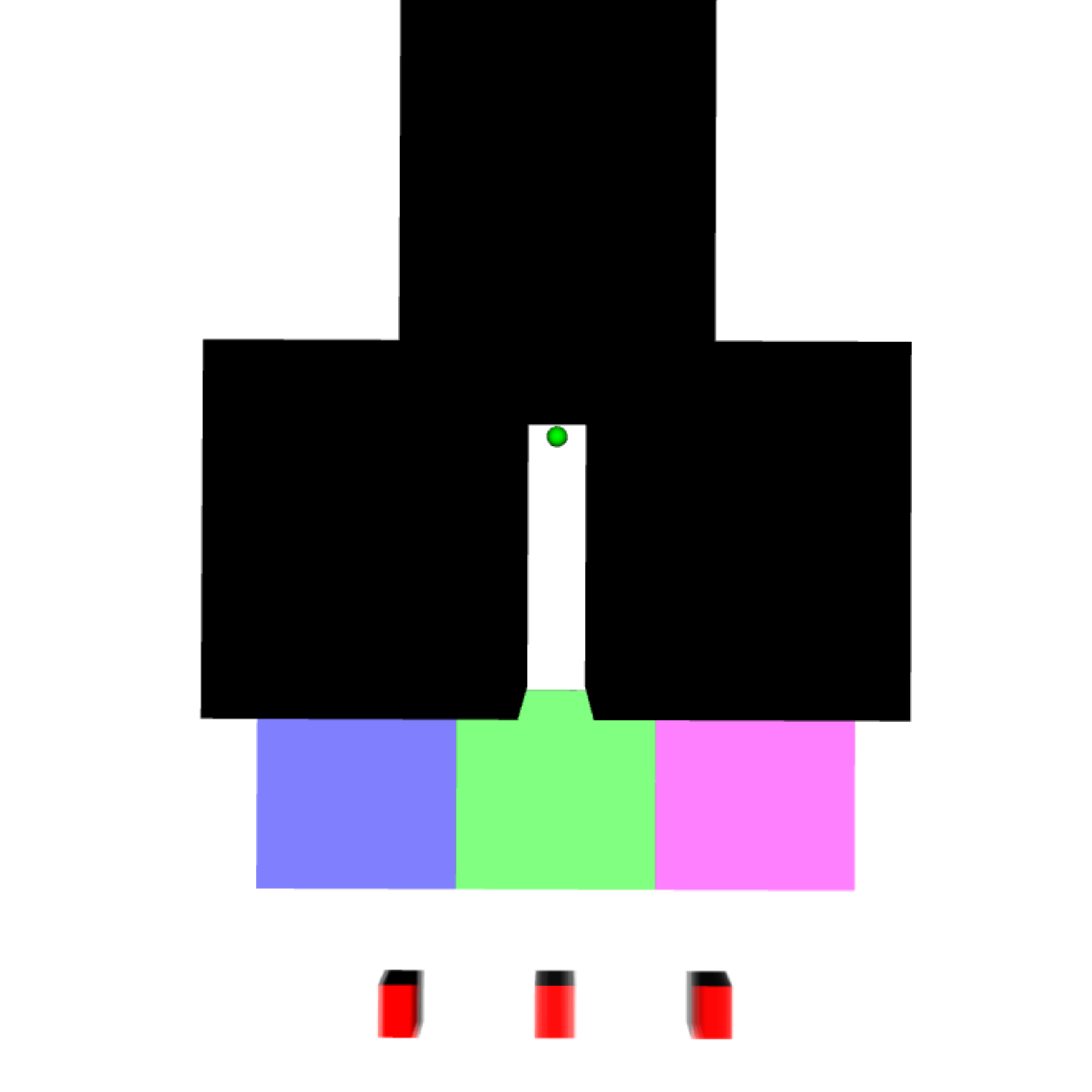} &
\includegraphics[height=0.25\textwidth]{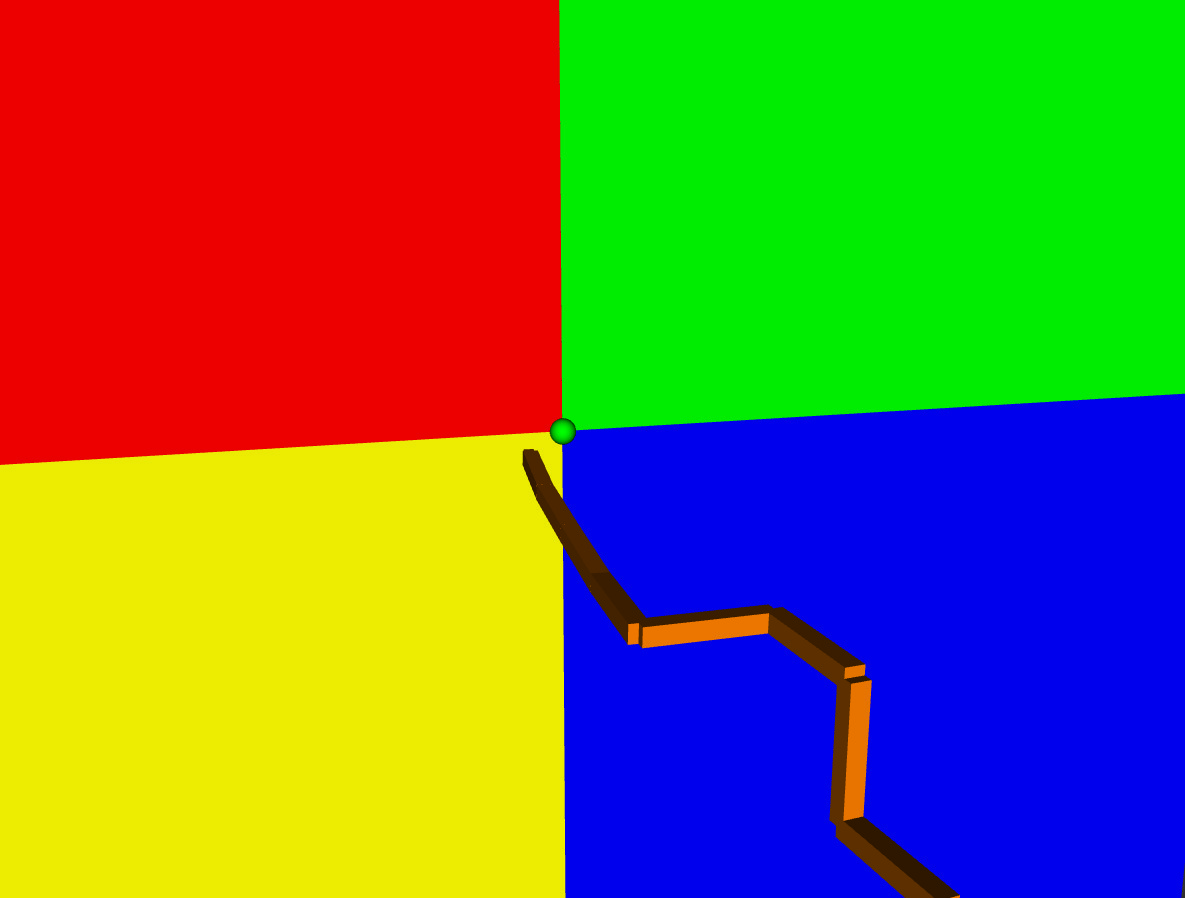} \\
(a) & (b) & (c)
\end{tabular}
\caption{Illustrations of (a) the Pushbox2D, (b) the Parking2D and (c) the SensorPlacement-8 problems. Goal regions are marked as green circles.}
\label{f:problemScenarios}
\end{figure}

\subsubsection{Pushbox}
Pushbox is a scalable motion planning problem motivated by air hockey. A disk-shaped robot has to push a disk-shape puck into a goal region (rewarded by $1,000$) by bumping into it, while avoiding collisions of itself and the opponent with a boundary region (penalized by $-500$). The robot can move freely in the environment by choosing a displacement vector. Upon bumping into the puck, it is pushed away and the robot's and puck's motions are affected by noise. The initial puck position is uncertain but the robot has access to a noisy bearing sensor to localize the puck. We consider two variants of the problem, \textbf{Pushbox2D} and \textbf{Pushbox3D} that differ in the dimensionality of the state and action spaces. For the \textbf{Pushbox2D} problem, the robot and the puck operate on a 2D-plane and the state space consists of the 2D-coordinates of the robot and the puck. The action space is $\actSpace\subset\mathbb{R}^2$ such that $\act\in\actSpace$ describes a displacement vector of the robot on the 2D-plane. \fref{f:problemScenarios}(a) illustrates the Pushbox2D problem. For \textbf{Pushbox3D}, both the robot and the puck move inside a 3D-environment and the action space is $\actSpace\subset\mathbb{R}^3$ where $\act\in\actSpace$ is a 3D-displacement vector. Additional details of the problem can be found in\ccite{seiler2015online}.


\subsubsection{Parking}
An autonomous vehicle with deterministic $2$nd-order dynamics\ccite{van2011lqg} operates in a 3D-environment populated by obstacles, shown in \fref{f:problemScenarios}(b). The goal of the vehicle is to safely navigate to a goal area located between the obstacles (rewarded by $100$) while avoiding collisions with the obstacles (black areas in \fref{f:problemScenarios}(b)), which is penalized by $-100$. We consider two variants of the problem, \textbf{Parking2D} and \textbf{Parking3D}. For \textbf{Parking2D}, the vehicle navigates on a 2D-plane and its state is a 4D-vector consisting of the $xy$-position of the vehicle on the plane, its orientation and its velocity. The vehicle is controlled via a steering wheel angle and acceleration, \ie, the action space is $\actSpace = \Omega\times\Phi$, where $\Omega$ is the continuous set of steering wheel angles and $\Phi$ is the continuous set of accelerations. There are three distinct areas in the environment, each consisting of a different type of terrain (colored areas in \fref{f:problemScenarios}(b)). Upon traversal, the vehicle receives an observation regarding the terrain type, which is only correct 70\% of the time due to sensor noise. Initially the vehicle starts near one of three possible starting locations (red areas in \fref{f:problemScenarios}(b)) with equal probability. The exact initial position of the vehicle along the horizontal $y$-axis is then drawn uniformly from $U[-0.175, 0.175]$ around the starting location. For \textbf{Parking3D} the vehicle operates in the full 3D space, and we have additional continuous state and action components that model the vehicles elevation and change in elevation respectively. The discount factor is $\gamma = 0.95$.

Two properties make this problem challenging: First is the multi-modal beliefs which require the vehicle to traverse the different terrains for a sufficient amount of time to localize itself before attempting to reach the goal. Second, due to the narrow passage that leads to the goal area, the robot requires precise motions in order to avoid collision with the obstacles. 

\subsubsection{Van Der Pol Tag}
Van Der Pol Tag (VDP-Tag) is a benchmark problem introduced in\ccite{sunberg2018online} in which an agent operates in a 2D-environment. The goal is to tag a moving target (rewarded by 100) whose motion is described by the Van Der Pol differential equation \commNew{and disturbed by Gaussian noise with standard deviation $\sigma=0.05$}. Initially, the position of the target is unknown. The agent travels at a constant speed but can pick its direction of travel at each step and whether to activate a costly range sensor (penalized by $5$), \ie, the action space is $\actSpace = [0, 2\pi) \times \{0, 1\}$, where the first component is the direction of travel and the second component is the activation/deactivation of the range sensor. The robot receives observations from its range sensor via 8 beams (\ie, $\obsSpace = \mathbb{R}^8$) that measure the agent's distance to the target if the target is within the beam's range. These measurements are more accurate when the range sensor is active. While the target moves freely in the environment, the agent's movements are restricted by four obstacles in the environment. More details of the VDP-Tag problem can be found in\ccite{sunberg2018online}. 
\commNew{For this problem, \solverAbbr needs to discretize the observation space, as detailed in the Appendix.}

\subsubsection{SensorPlacement}
We propose a scalable motion planning under uncertainty problem in which a $D$-DOF manipulator with $D$ revolute joints operates in muddy water inside a 3D environment. The robot is located in front of a marine structure, represented as four distinct walls, and its task is to mount a sensor at a particular goal area between the walls (rewarded with 1,000) while having imperfect information regarding its exact joint configuration. To localize itself, the robot's end-effector is equipped with a touch sensor. Upon touching a wall, it provides noise-free information regarding the touched wall. However, in order to avoid damage, the robot must avoid collisions (penalized by $-500$) between any of its other links and the walls. The state space of the robot consists of the set of joint-angles for each joint. The action space is $\actSpace \subset \mathbb{R}^D$, where $\act\in\actSpace$ is a vector of joint velocities. Due to underwater currents, the robot is subject to random control errors. Initially the robot is uncertain regarding its exact joint angle configuration. We assume that the initial joint angles are distributed uniformly according to $U[\theta_l, \theta_u]$, where $\theta_l = \theta_0 - h$ and $\theta_u = \theta_0 + h$, with $\theta_0$ corresponding to the configuration where all joint angles are zero, except for the second and third joint whose joint angles are $-1.57$ and $1.57$ respectively and $h = [0.1]^D$ (units are in radians). We consider four variants of the problem, denoted as SensorPlacement-$D$, with $D\in\{6, 8, 10, 12\}$, that differ in the degrees-of-freedom (number of revolute joints and thus the dimensionality of the action space) of the manipulator. \fref{f:problemScenarios}(c) illustrates the SensorPlacement-8 problem, where the colored areas represent the walls and the green sphere represents the goal area. The discount factor is $\gamma=0.95$. To successfully mount the sensor at the target location, the robot must use its touch sensor to carefully reduce uncertainty regarding its configuration. This is challenging, since a slight variation in the performed actions can quickly lead to collisions with the walls.
\subsection{Experimental Setup}
The purpose of our experiments is two-fold: First is to evaluate \solverAbbr and compare it with two state-of-the-art online POMDP solvers for continuous actions spaces, POMCPOW\ccite{sunberg2018online} and VOMCPOW\ccite{lim2020voronoi}. Note that in the original implementation of VOMCPOW and POMCPOW provided by the authors, the policy is recomputed after every planning step using a new search tree, whereas \solverAbbr applies ABT's\ccite{kurniawati2016online} strategy that re-uses the partial search tree (starting from the updated belief) constructed in the previous planning steps and improves the policy within the partial search tree. Therefore, we also tested modified versions of VOMCPOW and POMCPOW, called \textbf{VOMCPOW-I} and \textbf{POMCPOW-I}, where we follow \solverAbbr's strategy of re-using the partial search trees.

Second is to investigate the importance of the different components of \solverAbbr, specifically the Voronoi tree-based partitioning, the cell-diameter-aware exploration term in \eref{eq:u_value}, and the stochastic Bellman backups. For this purpose, we implemented the original \solverAbbr and three modifications. First is \textbf{\solverAbbr-R}, which replaces the Voronoi decomposition of \solverAbbr with a simple rectangular-based method: Each cell in the partition is a hyper-rectangle that is subdivided by cutting it in the middle along the longest side (with ties broken arbitrarily). The second variant is  \textbf{\solverAbbr(L=0)}, which is \solverAbbr where \eref{eq:u_value} reduces to the standard UCB1 bound. For the third variant, \textbf{\solverAbbr-MC}, we replace the stochastic Bellman backups described in \sref{ssec:backup} with simple Monte-Carlo backups 
as used by POMCPOW and VOMCPOW. Moreover, to test the effects of stochastic Bellman backups further, we implemented variants of the comparators, \textbf{VOMCPOW-B} and \textbf{POMCPOW-B}, which modify VOMCPOW-I and POMCPOW-I respectively to use stochastic Bellman backups instead of Monte Carlo backups. \commNew{Note that for the VDP-Tag problem, we did not test the variants of VOMCPOW and POMCPOW that re-use partial search trees, since each observation that the agent perceives from the environment leads to a new search tree due to the continuous observation space.}




To approximately determine the best parameters for each solver and problem, we ran a set of systematic preliminary trials by performing a grid-search over the parameter space. For each solver and problem, we used the best parameters and ran 1,000 simulation runs, with a fixed planning time of 1s CPU time for each solver and scenario. Each tested solver and the scenarios were implemented in C++ using the OPPT-framework\ccite{hoerger2018software}. All simulations were run single-threaded on an Intel Xeon Platinum 8274 CPU with 3.2GHz and 4GB of memory. 


\subsection{Results}

\begin{table*}
\centering
\caption{Average total discounted rewards and 95\% confidence intervals of all tested solvers on the Pushbox, Parking and VDP-Tag problems. The average is taken over 1000 simulation runs per solver and problem, with a planning time of 1s per step.}\label{t:results_1}
\vspace{-5pt}
\renewcommand{\arraystretch}{1.0}
\setlength{\tabcolsep}{2pt}
\begin{tabular}{l|*{5}{>{\hspace{0.55em}}rcl}}
& \multicolumn{3}{c}{Pushbox2D} & \multicolumn{3}{c}{Pushbox3D} &
	\multicolumn{3}{c}{Parking2D} & \multicolumn{3}{c}{Parking3D} & 
	\multicolumn{3}{c}{VDP-Tag} \\ \hline \hline
\solverAbbr & $351.6$ & $\hspace{-0.2em}\pm\hspace{-0.2em}$ & $10.0$ & $\mathbf{322.1}$ & $\hspace{-0.2em}\pm\hspace{-0.2em}$ &
$\mathbf{14.9}$ & $35.2$ & $\hspace{-0.2em}\pm\hspace{-0.2em}$ & $1.9$ & $\mathbf{32.6}$ & $\hspace{-0.2em}\pm\hspace{-0.2em}$ & ${3.5}$ & $24.3$ & $\hspace{-0.2em}\pm\hspace{-0.2em}$ & $1.2$ \\
\solverAbbr-R & $\mathbf{371.4}$ & $\hspace{-0.2em}\pm\hspace{-0.2em}$ & ${9.8}$ & $321.2$ & $\hspace{-0.2em}\pm\hspace{-0.2em}$ & $15.1$ &
$\mathbf{38.9}$ & $\hspace{-0.2em}\pm\hspace{-0.2em}$ & ${1.8}$ & $24.3$ & $\hspace{-0.2em}\pm\hspace{-0.2em}$ & $3.4$ & $24.5$ & $\hspace{-0.2em}\pm\hspace{-0.2em}$ & $1.2$  \\
\solverAbbr(L=0) & $340.8$ & $\hspace{-0.2em}\pm\hspace{-0.2em}$ & $14.7$ & $294.6$ & $\hspace{-0.2em}\pm\hspace{-0.2em}$ & $13.3$ & $29.2$ & $\hspace{-0.2em}\pm\hspace{-0.2em}$ & $3.5$ & $18.6$ & $\hspace{-0.2em}\pm\hspace{-0.2em}$ & $1.7$ & $24.2$ & $\hspace{-0.2em}\pm\hspace{-0.2em}$ & $1.1$\\
\solverAbbr-MC & $319.6$ & $\hspace{-0.2em}\pm\hspace{-0.2em}$ & $13.7$ & $311.0$ & $\hspace{-0.2em}\pm\hspace{-0.2em}$ & $16.2$ & $-3.2$ & $\hspace{-0.2em}\pm\hspace{-0.2em}$ & $1.8$ & $-14.7$ & $\hspace{-0.2em}\pm\hspace{-0.2em}$ & $0.5$ & $23.9$ & $\hspace{-0.2em}\pm\hspace{-0.2em}$ & $0.9$ \\ \hline
VOMCPOW-B & $322.9$ & $\hspace{-0.2em}\pm\hspace{-0.2em}$ & $12.1$ & $274.9$ & $\hspace{-0.2em}\pm\hspace{-0.2em}$ & $14.2$ & $28.2$ & $\hspace{-0.2em}\pm\hspace{-0.2em}$ & $1.8$ & $19.1$ & $\hspace{-0.2em}\pm\hspace{-0.2em}$ & $3.3$ & -\\
VOMCPOW-I & $316.0$ & $\hspace{-0.2em}\pm\hspace{-0.2em}$ & $12.3$ & $268.9$ & $\hspace{-0.2em}\pm\hspace{-0.2em}$ & $14.2$ & $-0.42$ & $\hspace{-0.2em}\pm\hspace{-0.2em}$ & $2.8$ & $-15.7$ & $\hspace{-0.2em}\pm\hspace{-0.2em}$ & $1.5$ &- \\
VOMCPOW & $129.8$ & $\hspace{-0.2em}\pm\hspace{-0.2em}$ & $13.3$ & $73.5$ & $\hspace{-0.2em}\pm\hspace{-0.2em}$ & $13.8$ & $-0.78$ & $\hspace{-0.2em}\pm\hspace{-0.2em}$ &
$2.8$ & $-18.4$ & $\hspace{-0.2em}\pm\hspace{-0.2em}$ & $1.4$ & $\mathbf{32.9}$ & $\hspace{-0.2em}\pm\hspace{-0.2em}$ & ${0.9}$ \\\hline
POMCPOW-B & $314.2$ & $\hspace{-0.2em}\pm\hspace{-0.2em}$ & $13.0$ & $245.7$ & $\hspace{-0.2em}\pm\hspace{-0.2em}$ & $14.1$ & $27.7$ & $\hspace{-0.2em}\pm\hspace{-0.2em}$ & $1.8$ & $8.8$ & $\hspace{-0.2em}\pm\hspace{-0.2em}$ & $2.6$ & -\\ 
POMCPOW-I & $270.6$ & $\hspace{-0.2em}\pm\hspace{-0.2em}$ & $18.9$ & $203.7$ & $\hspace{-0.2em}\pm\hspace{-0.2em}$ & $14.3$ & $-5.2$ & $\hspace{-0.2em}\pm\hspace{-0.2em}$ & $2.9$ & $-22.8$ & $\hspace{-0.2em}\pm\hspace{-0.2em}$ & $1.3$ & -\\ 
POMCPOW & $82.1$ & $\hspace{-0.2em}\pm\hspace{-0.2em}$ & $14.2$ & $3.6$ & $\hspace{-0.2em}\pm\hspace{-0.2em}$ & $12.9$ & $-5.1$ & $\hspace{-0.2em}\pm\hspace{-0.2em}$ & $3.0$ & $-25.7$ & $\hspace{-0.2em}\pm\hspace{-0.2em}$ & $1.4$ & $28.2$ & $\hspace{-0.2em}\pm\hspace{-0.2em}$ & $1.1$ \\ 
\end{tabular}
\end{table*}


\begin{figure}[!htb]
\centering
\small
\vspace{-5pt}
\includegraphics[width=0.8\textwidth]{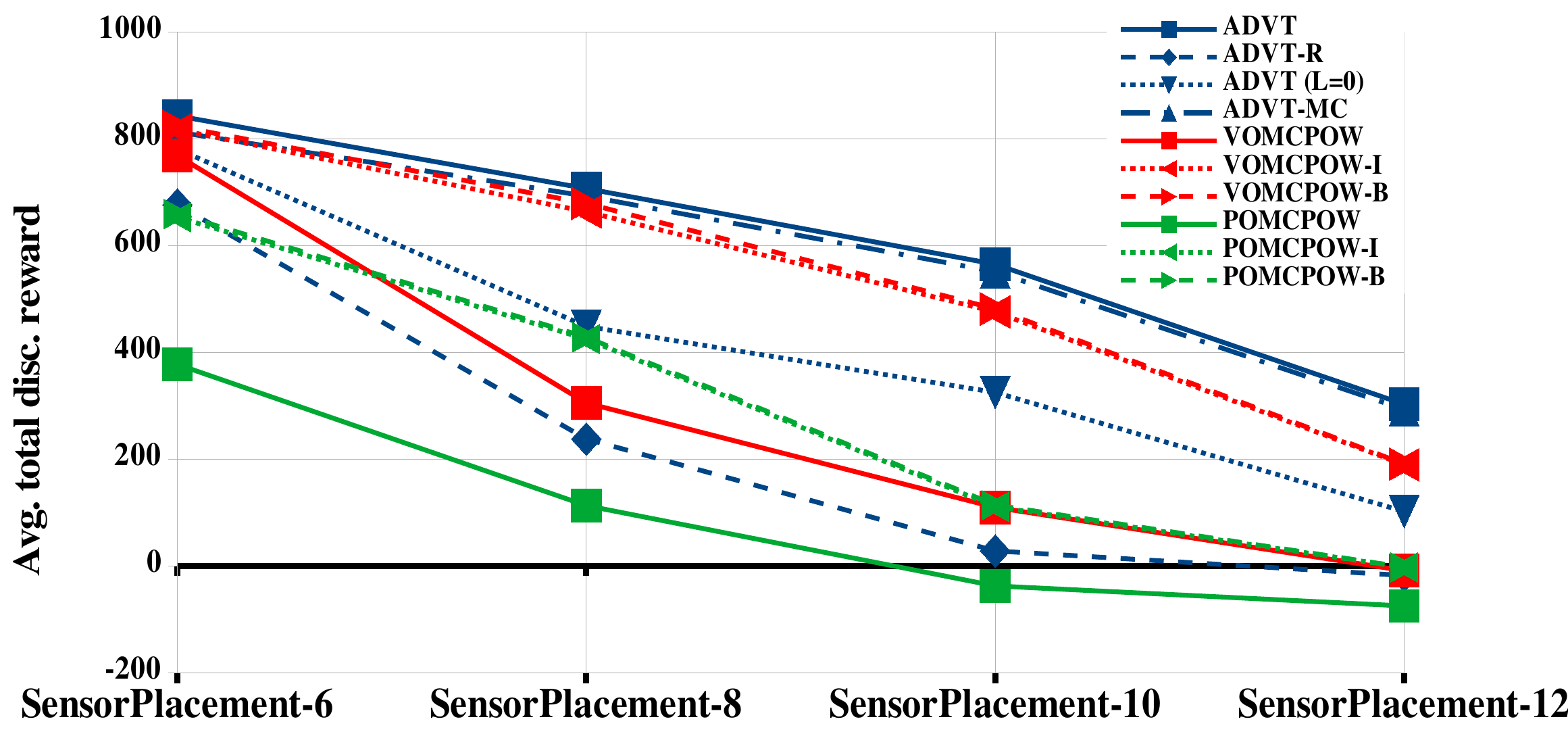}
\vspace{-5pt}
\caption{Average total discounted rewards of all tested solvers on the SensorPlacement problems. The average is taken over 1,000 simulation runs per solver and problem.}
\label{f:ResultsSensorPlacement}
\end{figure} 

\subsubsection{Comparison with State-of-the-Art Methods}\label{sssec:comparisonWithBaseline}
\tref{t:results_1} shows the average total discounted rewards of all tested solvers on the Pushbox, Parking and VDP-Tag problems, while \fref{f:ResultsSensorPlacement} shows the results for the SensorPlacement problems. Detailed results for the SensorPlacement problems \commNew{and results on the success rates of the tested solvers} are in the Appendix. 

\solverAbbr generally outperforms all other methods, except for VDP-Tag, where both POMCPOW and VOMCPOW perform better. The reason is that VDP-Tag's observation space is a continuous space. Both VOMCPOW and POMCPOW are designed for POMDPs with continuous state, action, and observation space, whilst \solverAbbr is \textbf{not} designed for continuous observation spaces. In this problem, \solverAbbr relies on discretizing the observation space, which is sub-optimal, considering the observation space is 8-dimensional, whereas VOMCPOW and POMCPOW apply Progressive Widening and weighted-particle representation of the beliefs in the search tree, which scales better to high-dimensional continuous observation spaces.
\commNew{Interestingly, in a VDP-Tag variant with a less noisy dynamics ($\sigma = 0.01$), \solverAbbr is only slightly worse than VOMCPOW and slightly better than POMCPOW, which supports \solverAbbr's effectiveness in handling continuous actions
(see the Appendix).} 
The results for the SensorPlacement problems indicate that \solverAbbr scales well to higher-dimensional action spaces.

\commNew{\solverAbbr performs well in terms of the success rate, too (Appendix 1.B). \solverAbbr maintains more than $90$\% success rate in the Pushbox, Parking and VDP-Tag problems and $>72$\% in the SensorPlacement-12 problem. VOMCPOW's and POMCPOW's success rate can drop as low as $\sim 30$\% and $12.5$\% in the Parking3D problem and to $<60$\% and $<40$\% in the SensorPlacement-12 problem respectively.}

Generally, VOMCPOW-I and POMCPOW-I perform much better than VOMCPOW and POMCPOW respectively, particularly in the Pushbox problems. These results (as well as those of \solverAbbr and all its variants) indicate the benefit of re-using partial search trees that were generated in previous planning steps instead of re-computing the policy at every planning step.

\subsubsection{Understanding the Effects of Different Components of \solverAbbr}\label{sssec:resultsAblationStudies} 
\paragraph{Effects of Voronoi-based partitioning for \solverAbbr.}
To understand the benefit of our Voronoi-based partitioning method, we compare the results of \solverAbbr with those of \solverAbbr-R. \tref{t:results_1} shows that \solverAbbr-R slightly outperforms \solverAbbr in the Pushbox2D, Parking2D and VDP-Tag problems, indicating that a rectangular-based partitioning works well for low-dimensional action spaces. However, \fref{f:ResultsSensorPlacement} shows that \solverAbbr-R is uncompetitive in the SensorPlacement problems as the dimensionality of the action space increases. For rectangular-based partitionings, the diameters of the cells can shrink very slowly in higher-dimensional action spaces. Additionally, the cell refinement method is independent of the spatial locations of the sampled actions. Both properties result in loose optimistic upper-confidence bounds of the $Q$-values, leading to excessive exploration of sub-optimal areas of the action space. For Voronoi trees, the geometries (and therefore the diameters) of the cells are much more adaptive to the spatial locations of the sampled actions, leading to more accurate optimistic upper-confidence bounds of the associated $Q$-values which avoids over-exploration of areas in the action space that already contain sufficiently many sampled actions. 

\paragraph{Effects of cell-size-aware optimistic upper-confidence bound.}
To investigate the importance of the component $L\ \mathrm{diam}(\cell)$ in the optimistic upper-confidence bound in \eref{eq:u_value}, let us compare \solverAbbr and \solverAbbr(L=0). The results in \tref{t:results_1} and \fref{f:ResultsSensorPlacement} indicate that the cell-diameter-aware component in the upper-confidence bound in \eref{eq:u_value} is important for \solverAbbr to perform well, particularly in the SensorPlacement problems. The reason is that in the early stages of planning, the partitions associated to the beliefs are still coarse, \ie, only a few candidate actions are considered per belief. If some of those candidate actions have small estimated $Q$-values, \solverAbbr(L=0) may discard large portions of the action space for a very long time, even if they potentially contain useful actions. The cell-diameter-aware bias term in \eref{eq:u_value} alleviates this issue by encouraging \solverAbbr to explore cells with large diameters. This is particularly important for problems with large action spaces such as the SensorPlacement problems.

\paragraph{Effects of Stochastic Bellman backups.}
To investigate the effects of this component, let us compare \solverAbbr with \solverAbbr-MC, as well as VOMCPOW and POMCPOW to VOMCPOW-B and POMCPOW-B, respectively. \tref{t:results_1} and \fref{f:ResultsSensorPlacement} reveal that the solvers that use stochastic Bellman backups (\solverAbbr, \solverAbbr-R, VOMCPOW-B and POMCPOW-B) perform significantly better, particularly in the Parking2D and Parking3D problems. The reason is that in both problems good rewards are sparse, particularly for beliefs where the vehicle is located between the walls and slight deviations from the optimal actions can lead to collisions. The stochastic Bellman backups help to focus the search on more promising regions of the action space.

\section{Conclusion}

We propose a new sampling-based online POMDP solver, called \solverAbbr, that scales well to POMDPs with high-dimensional continuous action spaces. 
Our solver builds on a number of works that uses adaptive discretization of
the action space, and introduces a more effective adaptive discretization method
that uses novel ideas:
A Voronoi tree based adaptive hierarchical discretization of the action space, 
a novel cell-size aware refinement rule, 
and a cell-size aware upper-confidence bound.
\solverAbbr shows strong empirical results against state-of-the-art algorithms
on several challenging benchmarks. We hope this work further expands the applicability of general-purpose POMDP solvers. In future works we are planning to extend \solverAbbr to handle continuous observation spaces as well, which would allow us to tackle even more challenging POMDP problems. 

\subsubsection{Acknowledgements} 
We thank Jerzy Filar for many helpful discussions. 
This work is partially supported by the Australian Research Coucil (ARC)
Discovery Project 200101049. 

%
%
%
\bibliographystyle{splncs03}

\input{main.bbl}
\newpage
\begin{subappendices}

\input{appendix.tex}
\end{subappendices}

\end{document}

%% file: shortcut.tex
\newcommand{\belSpace}{\ensuremath{\mathcal{B}}\xspace}
\newcommand{\bel}{\ensuremath{b}\xspace}

\newcommand{\stSpace}{\ensuremath{\mathcal{S}}\xspace}

\newcommand{\st}{\ensuremath{s}\xspace}
\newcommand{\stp}{\ensuremath{s'}\xspace}

\newcommand{\actSpace}{\ensuremath{\mathcal{A}}\xspace}
\newcommand{\act}{\ensuremath{a}\xspace}

\newcommand{\obsSpace}{\ensuremath{\mathcal{O}}\xspace}
\newcommand{\obs}{\ensuremath{o}\xspace}

\newcommand{\transF}{\ensuremath{T}\xspace}
\newcommand{\obsF}{\ensuremath{Z}\xspace}

\newcommand{\rewFunc}{\ensuremath{R}\xspace}

\newcommand{\pol}{\ensuremath{\pi}\xspace}
\newcommand{\optPol}{\ensuremath{\pi^*}\xspace}

\newcommand{\belTree}{\ensuremath{\mathcal{T}}\xspace}
\newcommand{\VT}{\ensuremath{\mathcal{H}}\xspace}

\newcommand{\episode}{\ensuremath{e}\xspace}

\newcommand{\cell}{\ensuremath{P}}

\newcommand{\ccite}[1]{~\cite{#1}}

\newcommand{\sref}{Section~\ref}

\newcommand{\aref}[1]{Algorithm~\ref{#1}}
\newcommand{\eref}[1]{eq.(\ref{#1})}
\newcommand{\fref}[1]{Figure~\ref{#1}}

\newcommand{\tref}[1]{Table~\ref{#1}}

\newcommand{\solver}{{\bf A}daptive {\bf D}iscretization using {\bf V}oronoi {\bf T}rees\xspace}
\newcommand{\solverAbbr}{ADVT\xspace}

\newcommand{\ie}{i.e.\xspace}

\newcommand{\expect}[1]{\ensuremath{\mathbb{E}[#1]}\xspace}
\newcommand{\del}[1]{\mathrm{d}{#1}}

\DeclareMathOperator{\reals}{\mathbb{R}}

\newcommand{\pomdpTuple}{\ensuremath{\mathcal{P}}\xspace}

\newcommand{\commNew}[1]{{#1}}

%% file: appendix.tex
\section{Detailed Experimental Results for the SensorPlacement Problems}

\begin{table*}
\centering
\caption{Average total discounted rewards and 95\% confidence intervals of all tested solvers on the SensorPlacement problems. The average is taken over 1000 simulation runs per solver and problem, with a planning time of 1s per step.}\label{t:results_2}
\scalebox{0.88}{
\begin{tabular}{l|cccc}
& SensorPlacement-6 & SensorPlacement-8 & SensorPlacement-10 & SensorPlacement-12\\ \hline \hline
\solverAbbr & \textbf{842.8 $\pm$ 9.5} & \textbf{706.8 $\pm$ 17.5} & \textbf{565.1 $\pm$ 21.7} & \textbf{303.0 $\pm$ 19.8} \\
\solverAbbr-R & 676.3 $\pm$ 19.7 & 238.1 $\pm$ 33.5 & 28.7 $\pm$ 18.4 & -17.3 $\pm$ 7.4 \\
\solverAbbr ($L=0$) & 780.4 $\pm$ 12.6 & 448.8 $\pm$ 15.9 & 325.3 $\pm$ 16.2 & 102.5 $\pm$ 7.4 \\
\solverAbbr-MC & 812.6 $\pm$ 11.4 & 692.7 $\pm$ 17.7 & 551.2 $\pm$ 18.3 & 293.5 $\pm$ 19.6 \\ \hline
VOMCPOW-B & 823.4 $\pm$ 15.1& 679.1 $\pm$ 17.9& 481.6 $\pm$ 22.2& 191.3 $\pm$ 17.6 \\ 
VOMCPOW-I & 817.2 $\pm$ 15.8& 663.4 $\pm$ 18.6& 476.0 $\pm$ 22.7& 189.9 $\pm$ 18.0 \\
VOMCPOW & 768.5 $\pm$ 16.4 & 305.6 $\pm$ 25.8 & 110.1 $\pm$ 24.5 & -8.2 $\pm$ 13.2\\ \hline
POMCPOW-B & 659.3 $\pm$ 17.2 & 428.7 $\pm$ 21.5 & 114.6 $\pm$ 16.6& -1.9 $\pm$ 6.6 \\
POMCPOW-I & 653.2 $\pm$ 17.3 & 425.2 $\pm$ 21.8 & 111.3 $\pm$ 16.8& -2.1 $\pm$ 6.8 \\
POMCPOW & 377.6 $\pm$ 23.5 & 113.4 $\pm$ 24.2 & -36.8 $\pm$ 11.3 & -74.3 $\pm$ 12.9 \\
\end{tabular} 
}
\end{table*}

\section{Success Rates}
\begin{table*}
\centering
\caption{Success rates of all tested solvers on the Pushbox, Parking and VDP-Tag problems. The success rate is with respect to 1,000 simulation per solver and problem, with a planning time of 1s per step.}
\label{t:results_success_rate}
\vspace{-5pt}
\renewcommand{\arraystretch}{1.0}
\setlength{\tabcolsep}{2pt}
\begin{tabular}{l|*{5}{>{\hspace{0.55em}}rcl}}
& \multicolumn{3}{c}{Pushbox2D} & \multicolumn{3}{c}{Pushbox3D} &
	\multicolumn{3}{c}{Parking2D} & \multicolumn{3}{c}{Parking3D} & 
	\multicolumn{3}{c}{VDP-Tag} \\ \hline \hline
\solverAbbr & $0.985$ &  &  & $0.969$ &  &  & $0.912$ &  &  & $0.916$ &  &  & $0.941$ & &  \\
\solverAbbr-R & $0.987$ &  &  & $0.968$ &  &  & $0.943$ &  &  & $0.906$ & &  & $0.945$ &  &  \\
\solverAbbr(L=0) & $0.966$ &  &  & $0.965$ &  &  & $0.857$ &  &  & $0.898$ &  &  & $0.935$ &  & \\
\solverAbbr-MC & $0.989$ &  &  & $0.972$ &  &  & $0.417$ &  &  & $0.337$ &  & & $0.892$ &  & \\ \hline
VOMCPOW-B & $0.985$ & &  & $0.970$ &  &  & $0.885$ &  &  & $0.886$ &  &  & -\\
VOMCPOW-I & $0.975$ &  & & $0.939$ &  &  & $0.597$ &  &  & $0.314$ &  &  &- \\
VOMCPOW & $0.754$ &  &  & $0.815$ &  &  & $0.512$ &  &  & $0.297$ & &  & $0.987$ &  &  \\\hline
POMCPOW-B & $0.974$ &  &  & $0.953$ &  &  & $0.853$ &  &  & $0.534$ &  &  & -\\ 
POMCPOW-I & $0.963$ &  &  & $0.969$ &  &  & $0.409$ &  &  & $0.122$ &  &  & -\\ 
POMCPOW & $0.712$ &  &  & $0.692$ &  & & $0.401$ & & & $0.125$ &  &  & $0.979$ & & \\ 
\end{tabular}
\end{table*}

\begin{table*}
\centering
\caption{Success rates of all tested solvers on the SensorPlacement problems. The success rate is with respect to 1,000 simulation per solver and problem, with a planning time of 1s per step.}
\label{t:results_success_rate_2}
\scalebox{0.88}{
\begin{tabular}{l|cccc}
& SensorPlacement-6 & SensorPlacement-8 & SensorPlacement-10 & SensorPlacement-12\\ \hline \hline
\solverAbbr & 0.981 & 0.962 & 0.834 & 0.724 \\
\solverAbbr-R & 0.832 & 0.692 & 0.756 & 0.557 \\
\solverAbbr ($L=0$) & 0.937 & 0.726 & 0.791 & 0.601 \\
\solverAbbr-MC & 0.964 & 0.959 & 0.828 & 0.719 \\ \hline
VOMCPOW-B & 0.979 & 0.951 & 0.807 & 0.703 \\ 
VOMCPOW-I & 0.967 & 0.891 & 0.803& 0.698 \\
VOMCPOW & 0.923 & 0.721 & 0.645 & 0.583\\ \hline
POMCPOW-B & 0.829 & 0.794 & 0.646 & 0.575 \\
POMCPOW-I & 0.826 & 0.781 & 0.657 & 0.578 \\
POMCPOW & 0.738 & 0.636 & 0.519 & 0.321 \\
\end{tabular} 
}
\end{table*}

In addition to the average total discounted rewards in the main document, we also report the success rate of each solver in each problem scenario in \tref{t:results_success_rate} and \tref{t:results_success_rate_2}. For the Pushbox problems, a run is considered successful if the robot pushes the opponent into the goal region, while avoiding collisions of itself and the opponent with the boundary region. For the Parking problems, a run is successful if the vehicle reaches the goal area. For the VDP-Tag problem a run is successful if the opponent is being tagged and for the SensorPlacement problems, a run is successful if the end-effector reaches the sensor mounting location. In all problems, the task must be completed within planning steps 50 steps, otherwise problem terminates and the run is considered unsuccessful. 

Generally the success rates are closely correlated to the average total discounted rewards achieved by each solver in the problem scenarios. The results in \tref{t:results_success_rate_2} further indicate that \solverAbbr scales better to higher-dimensional action spaces compared to the baselines. However, for the SensorPlacement12 problem, the success rates are relatively low, even for \solverAbbr. Thus, for such problems, developing methods that scale even better to high-dimensional action spaces remains a fruitful avenue for future research.

\section{Observation Discretization Method for the VDP-Tag Problem}\label{sec:VDP_discretization}
Since the observation space in the VDP-Tag problem is continuous, \ie, $\obsSpace = \mathbb{R}^8$, ADVT requires a method to discretize the observations. To this end, we use a simple distance-based discretization: Suppose a sampled episode selects action $\act\in\actSpace(\bel)$ at belief \bel and perceives an observation $\obs_i\in\obsSpace$. We then compute the Euclidean distance between $\obs_i$ and each observation corresponding to the outgoing edges $(\act, \obs)\in\belTree$ that descend \bel. If there is an observation $\obs_k$ corresponding to an outgoing edge for which the Euclidean distance to $\obs_i$ yields a value smaller than a threshold $\delta$ (in our experiments we use $\delta = 25$), we continue the search from child node $\bel'$ of \bel via edge $(\act, \obs_k)$. Otherwise, we add a new child node to \bel via edge $(\act, \obs_i)$.

\section{Results for the VDP-Tag Problem With Smaller Transition Errors}

\begin{table*}
\centering
\caption{Average total discounted rewards with 95\% confidence intervals and success rates of all tested solvers on the VDP-Tag with smaller transition errors ($\sigma = 0.01$). The average is taken over 1000 simulation runs per solver and problem, with a planning time of 1s per step.}
\label{t:results_vdp2}
\setlength{\tabcolsep}{10pt}
\begin{tabular}{l|*{3}{>{\hspace{2.55em}}rcl}}
& \multicolumn{3}{c}{Avg. total discounted reward} & \multicolumn{3}{c}{Success rate} \\ \hline \hline
\solverAbbr & $31.5$ & \hspace{-18pt}$\pm$ & \hspace{-18pt}$0.9$ & \multicolumn{3}{c}{$0.991$} \\
\solverAbbr-R & $31.7$ & \hspace{-18pt}$\pm$ & \hspace{-18pt}$0.9$ & \multicolumn{3}{c}{$0.986$} \\
\solverAbbr L(=0) & $30.1$ & \hspace{-18pt}$\pm$ & \hspace{-18pt}$1.0$ & \multicolumn{3}{c}{$0.989$} \\
\solverAbbr-MC & $30.9$ & \hspace{-18pt}$\pm$ & \hspace{-18pt}$0.8$ & \multicolumn{3}{c}{$0.984$} \\
VOMCPOW & $\mathbf{34.1}$ & \hspace{-18pt}$\mathbf{\pm}$ & \hspace{-18pt}$\mathbf{0.8}$ & \multicolumn{3}{c}{$\mathbf{0.998}$} \\
POMCPOW & $29.1$ & \hspace{-18pt}$\pm$ & \hspace{-18pt}$0.8$ & \multicolumn{3}{c}{$0.990$} \\
\end{tabular} 
\end{table*}

We additionally tested \solverAbbr as well as VOMCPOW and POMCPOW on a variant of the VDP-Tag problem with smaller transition errors, \ie, the position of the target is disturbed by Gaussian noise with standard deviation $\sigma = 0.01$ instead of $\sigma = 0.05$. The results are shown in \tref{t:results_vdp2}.

It can be seen that for the variant of the problem with $\sigma = 0.01$, \solverAbbr is competitive with POMCPOW. For $\sigma = 0.05$ the uncertainty with respect to the position of the target is large and therefore the agent must carefully decide when to activate its range sensor in order to reduce uncertainty. To achieve this, the solvers require more accurate belief representations in the search tree, which is challenging for \solverAbbr, as it relies on discretizing the continuous observation space. On the other hand, VOMCPOW and POMCPOW use Progressive Widening in the observation space, combined with a weighted-particle representation of the beliefs in the search trees, which helps them to perform well.  

For $\sigma = 0.01$, the uncertainty with respect to the position of the target is much smaller, and therefore the solvers are less reliant on accurate belief representations in the search tree, which benefits \solverAbbr. This suggests that our method works well for problems with small belief uncertainties, even if the observation space is continuous. To handle larger uncertainties, we require a better method to handle continuous observation spaces, such as progressive widening and explicit belief representations as used in VOMCPOW and POMCPOW. We are planning to explore this in future works.